# LexGen: Domain-aware Multilingual Lexicon Generation

**Ayush Maheshwari**[* ♡ †], **Atul Kumar Singh**[* ♠♣], **Karthika N J**[♠♣],
**Krishnakant Bhatt**[♠], **Preethi Jyothi**[♠], **Ganesh Ramakrishnan**[♠♣]
[♠]Indian Institute of Technology Bombay [♣]BharatGen
[♡]NVIDIA
{ayusham, aksingh}@cse.iitb.ac.in

## Abstract

Lexicon or dictionary generation across domains has the potential for societal impact, as it can potentially enhance information accessibility for a diverse user base while preserving language identity. Prior work in the field primarily focuses on bilingual lexical induction, which deals with word alignments using mapping or corpora-based approaches. However, these approaches do not cater to domain-specific lexicon generation that consists of domain-specific terminology. This task becomes particularly important in specialized medical, engineering, and other technical domains, owing to the highly infrequent usage of the terms and scarcity of data involving domain-specific terms especially for low/mid-resource languages. In this paper, we propose a new model to generate dictionary words for 6 Indian languages in the multi-domain setting. Our model consists of domain-specific and domain-generic layers that encode information, and these layers are invoked via a learnable routing technique. We also release a new benchmark dataset consisting of >75K translation pairs across 6 Indian languages spanning 8 diverse domains. We conduct both zero-shot and few-shot experiments across multiple domains to show the efficacy of our proposed model in generalizing to unseen domains and unseen languages. Additionally, we also perform a post-hoc human evaluation on unseen languages. The source code and dataset is present at `https://github.com/Atulkmrsingh/lexgen`.

## 1 Introduction

Lexicons are a vital resource for several natural language processing tasks, providing information about word meanings, semantic relationships, and linguistic properties. Dictionaries[1] play a crucial role in facilitating knowledge dissemination within specialized subjects or domains. While certain semantic traits may be shared across domains, substantial heterogeneity exists in lexicon usage across subject areas.

The task of dictionary generation is less explored compared to dictionary induction approaches. Bilingual lexicon induction (BLI) approaches such as (Li et al., 2022; Tian et al., 2022; Jawanpuria et al., 2020; Lample et al., 2017) induce word or phrase translations by aligning independently trained word embeddings in two languages in a shared embedding space. BLI methods often rely on pre-trained embeddings, supervised word-pairs and co-occurrence patterns within a limited window of text. However, multi-word lexicon induction is a challenge because induction methods focus on alignment of word pairs in both supervised and unsupervised setting. Additionally, these methods do not focus on enhancing the features of domain-specific words.

Meanwhile, multi-lingual neural machine translation (MNMT) (Arivazhagan et al., 2019; Zhang et al., 2020b) approaches have achieved remarkable success in generating accurate and contextually relevant translations, given the availability of a huge amount of annotated parallel corpora. However, a parallel corpus is often limited, particularly in the domains of engineering, medicine, *etc*. As a result, pre-trained MNMT models have limited capacity for generating new words, especially unseen domains. Unlike induction methods, LexGen leverages the co-occurrence knowledge encoded by the pre-trained translation model and is trained in a domain-aware mechanism to produce translations.

For low-resource languages, limited linguistic resources and sparse data pose challenges in the development of comprehensive and accurate dictionaries. In this paper, we propose LexGen, an approach to address the task of generating domain-specific dictionaries under a limited supervision setting. Our objective is to generate translations in zero-shot and few-shot settings from English to six Indic languages, *viz.*, Hindi, Kannada, Gujarati, Marathi, Odia, and Tamil. Additionally, we explore translations across several diverse domains, encompassing both technical and non-technical subject areas.

[*] Authors contributing equally.
[†] Work done while pursuing PhD at IIT Bombay.
[1] Throughout the paper, we use the terms 'lexicon' and 'dictionary' interchangeably.

In this paper, we perform extensive experiments on 6 Indic languages and 8 diverse domains under different scenarios (Section 4). These settings include scenarios where the training and test data are: (i) **in**-**d**omain and in the **s**ame **t**arget language (IDST) (*c.f.*, Section 4), (ii) **in**-**d**omain but **d**ifferent **t**arget languages (IDDT) (zero-shot) (*c.f.*, Section 4.2), and (iii) **d**ifferent **d**omains but the **s**ame **t**arget language (DDST) (zero-shot) (*c.f.*, Section 5.3). In Figure 1, we present the settings with illustrative examples. We observed an average improvement of 2-5% compared to baselines in all languages and domains. We attribute this improvement to the domain-sensitive architecture of LexGen and to the fact that Indian languages tend to share common stems while differing in language-specific affixes (Masica, 1991; Cardona and Jain, 2007). Finally, in Section 5.5, we report post-hoc human evaluations on the predictions generated by the best models on the same domain but with different target languages. Our contributions can be summarised as follows:

1. We propose an end-to-end framework LexGen, which allows dynamic routing (*c.f.*, Section 3.1) of information between domain-specific and shared paths. Inspired by Zhang et al. (2020a), we incorporate a token-level activation gating mechanism in our Domain Routing (DR) layer, enabling the selective flow of information to domain-specific or shared networks. From extensive experiments and results, we find that our model outperforms the baselines, including the base Transformer model on which the framework is built and other large multilingual models.

2. For further research and benchmarking, we will provide our curated datasets consisting of more than 75K dictionary translation pairs for public use. These test sets serve as valuable resources for evaluating and comparing both domain-aware dictionary generation and induction methods across different languages and domains.

## 2 Related Work

Bilingual Lexicon Induction (BLI) methods (Bafna et al., 2024; Tian et al., 2022; Patra et al., 2019; Lample et al., 2018), aims to leverage pre-trained monolingual word embeddings across different languages and aligning them into a shared space through linear transformations. Unsupervised BLI approaches (Lample et al., 2018; Artetxe et al., 2018; Mohiuddin and Joty, 2019; Ren et al., 2020; Han et al., 2023) attempts to align word embeddings through generative adversarial network-based unsupervised learning to induce word translations. While semi-supervised BLI approaches (Patra et al., 2019; Mohiuddin et al., 2020; Zhao et al., 2020) leverage small seed dictionaries by emphasizing two-way interaction between the unsupervised lexicon induction loss and supervised loss. On the other hand, dictionary generation approaches such as pivot-based methods (Varga and Yokoyama, 2009) and graph-based methods (Goel et al., 2022; Bestgen, 2022) rely on the existence of a pivot language to discover translation between two languages, building on the principle of transitive correspondence.

Recently, RAPO (Tian et al., 2022) formulates BLI as a ranking problem and creates personalized mappings for words leveraging set cluster similarities. Similarly, BLICEr (Li et al., 2022), enhances BLI by fine-tuning a cross-encoder model on the contrastive word pairs induced from the pre-trained model. Li et al. (2023) examines the potential of large language models (LLMs) for the task of BLI and investigates zero-shot and few-shot in-context prompting. However, their approach is limited by the language capability of LLMs. In contrast, our approach leverages pre-trained translation models for the task of domain-aware lexicon induction. Further, we also compare LexGen with fine-tuning and in-context learning with LLMs.

## 3 The LexGen Framework

Given a source phrase $X$ in source language $i$ and corresponding target translation $Y$ in target language $j$, we reuse multi-lingual NMT models for source-to-target translation. In multi-lingual NMT models, the source phrase is appended with a target language tag such that $\hat{X} = \{\text{lang}, X\}$. We use a pre-trained Transformer-based (Vaswani et al., 2017) translation model based on the encoder-decoder architecture consisting of multiple encoder and decoder blocks. The encoder is a stack of $L$ identical blocks with each block $l$ containing a self-attention network (SAN) and a feed-forward network (FFN) layer. Each decoder block contains an additional cross-attention network (CAN) layer between SAN and FFN to learn the alignment between source and target tokens. Within each encoder and decoder block, the input to the current layer, $z_l$, and output from the current layer, $f(z_l)$, is combined using layer-normalization (LN) (Ba et al., 2016) to smoothen information flow and prevent gradient vanishing and explosion.

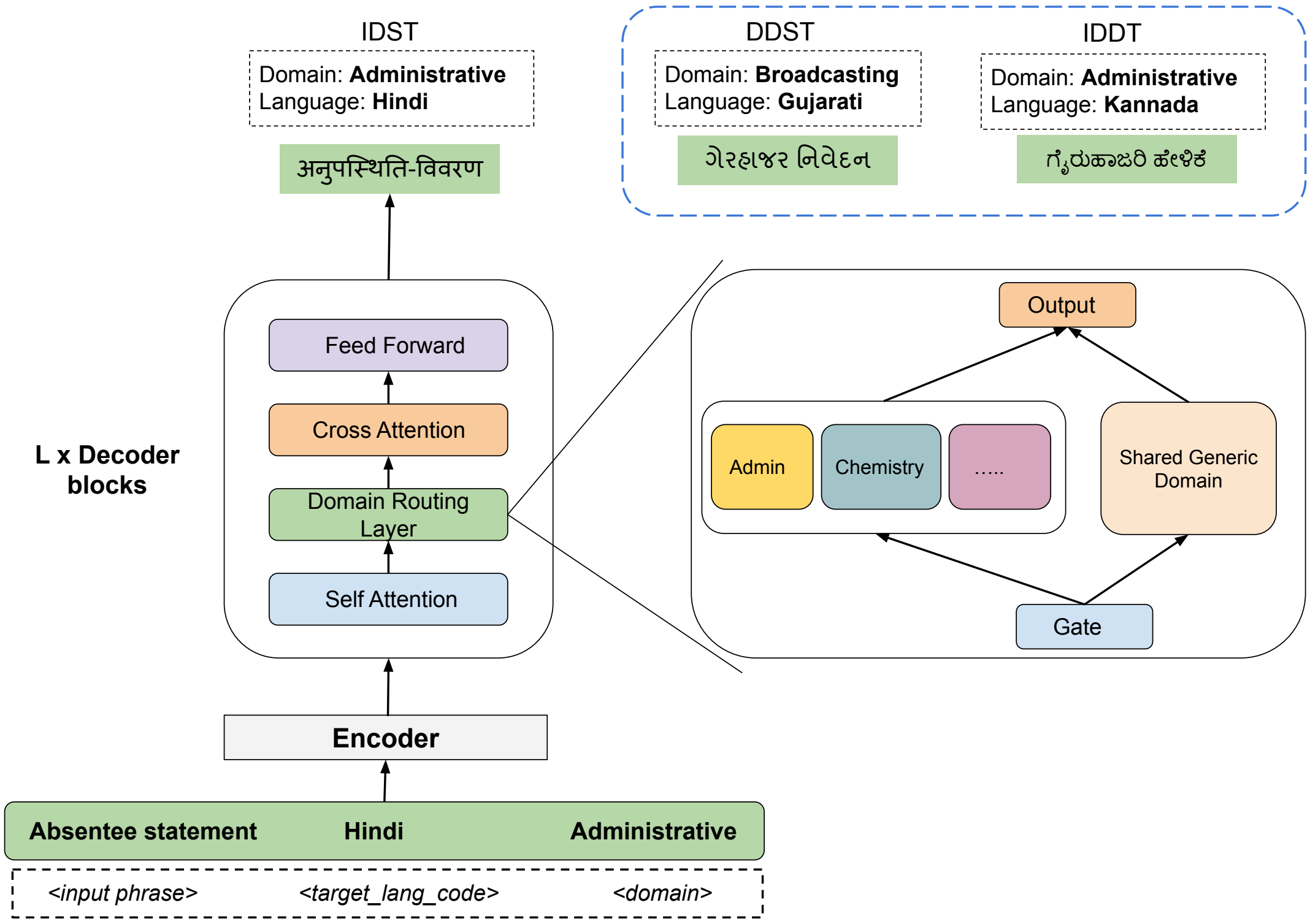

Figure 1: Model architecture of the LexGen framework. For an input English phrase, we show sample outputs for three settings we adopt in our experiments. The dashed blue boxes demonstrate the output for IDDT and DDST experimental settings. We introduce a domain routing layer after every self-attention layer in each decoder block. The gating layer learns to route each decoder input through either a domain-specific layer or a shared generic layer.

### 3.1 Domain Routing (DR)

Previous works have employed adapters and language routing mechanisms to enhance the expressiveness of models and enable them to capture domain-specific characteristics. Building upon the insights from routing and domain-adapter approaches (Liu et al., 2022; Vu et al., 2022; Zhang et al., 2020a), our method incorporates a domain routing (DR) layer. This layer allows us to capture domain-specific characteristics and facilitate information sharing across domains. By leveraging pre-trained NMT models, we mitigate the challenge of language expressivity since these models have demonstrated the ability to effectively capture the alignment between different languages, especially for small-length sentences, in training corpora. This allows us to focus on domain-specific aspects without compromising language understanding and translation quality. Furthermore, the DR layer improves the ability of the NMT model to capture domain-specific linguistic units and enhance the accuracy and specificity of the generated translations. We show several examples of generated translations in the Appendix (Table 9).

As depicted in Figure 1, we use a pre-trained multilingual model to initialize the parameters of our encoder and decoder layers. In each decoder block, we infuse a domain routing (DR) layer after each self-attention layer. The DR layer parameters are shared across decoder blocks to avoid over-fitting on a (possibly) small number of training instances. The intuition behind the DR layer is to discriminate between domain-specific words and generic words. DR learns a real-valued gating value $g(.)$ for each input token based on the output of the previous layer. In addition, the DR layer controls the flow of the output from the previous sub-layer via learnable domain-specific or domain-generic channels.

$$\begin{aligned} \mathrm{DR}(f(z_l)) &= g(z_l) W_{\text{dom}} f(z_l) \\ &+ (1 - g(z_l)) W_{\text{shared}} f(z_l) \end{aligned} \quad (1)$$

where $W_{\text{dom}}$ is a weight matrix designed to capture domain-specific information while $W_{\text{shared}}$ is a shared weight matrix across domains. Note that $W_{\text{dom}}$ and $W_{\text{shared}}$ are shared across all languages and we rely on multilingual pre-trained models to capture language specific nuances. Intuitively, these matrices effectively encode and learn the distinct characteristics of each domain as well as the shared knowledge that is relevant across all domains. The input will traverse through different gates of the DR layer and train $W_{\text{dom}}$ differently for each domain, irrespective of the language.

Inspired by gating-based approaches (Chen et al., 2020; Maheshwari et al., 2024; Zhang et al., 2020a), we additionally learn a binary gate value to control the flow of information to the domain-specific ($W_{\text{dom}}$) and shared ($W_{\text{shared}}$) weight matrices[2]. During training, we parameterize the gate $g(.)$ with the following two-layer feed-forward network:

$$g(z_l) = \sigma(\mathrm{Relu}(z_l W_1 + b) W_2)$$

where Relu(.) is the ReLU function, $\sigma(.)$ is the logistic sigmoid function, $W_1$ and $W_2$ are linear layers with trainable parameters. The domain-specific routing (DR) layer learns a gating value $g(.)$ for each token based on the output $z_l$ from the previous layer. Note that we fine-tune all the parameters of the pre-trained model and DR layer during training.

Further, we conduct several ablation experiments in which we a) remove the DR layer, b) place the DR layer after SAN, and c) place the DR layer after CAN. We found that placing the DR layer after the SAN provides the best performance. across various domains and languages (*c.f.* Table 5). Additionally, we also attempted to add Gaussian noise to sigmoid gating functions to discretize the flow of information through the gates (Bapna et al., 2020; Chiu and Raffel, 2018). However, this did not yield improvements in our results, hence we omitted its use in subsequent experiments.

[2] $W_1$ has the dimension 1536 x 256 and $W_2$ has the dimension 256 x 1.

## 4 Experimental Setup

### 4.1 In-domain same-target language (IDST)

During the training phase, our dataset consists of dictionary pairs in the domain $m$ for the source language $i$ and the corresponding target language $j$. In this setting, the test set consists of instances from the same domain and the same source and target languages. Training is performed on dictionary pairs from English to 6 Indic languages for the administrative, biotechnology, and chemistry domains. We randomly choose 10% of instances for each domain and target language as our test set. The purpose of this experiment is to assess the effectiveness and efficiency of our methodology in generating accurate translations within these specific domains and language pairs.

### 4.2 In-domain different-target language (IDDT) (Zero-shot)

In this experiment, we aim to assess the language generalization capabilities of our approach. The training phase involved English-Hindi, Tamil, and Gujarati language pairs from the domains of biotechnology, administration, and chemistry. We do a cross-lingual evaluation on the same domain, but from English to Kannada, Marathi, and Odia language pairs. By evaluating our approach's performance in this transfer learning scenario, we gain insights into its ability to effectively handle new language pairs within familiar domains. The test set corresponds to the same instances as in the IDST setting but on unseen target languages.

### 4.3 Evaluation Metrics

Indic languages (including the six languages we experiment with) are morphologically rich and given the shorter word length of the target words as well as the predictions, we use ChrF++ metric (Popović, 2017) for the dictionary generation task. ChrF++ exhibits better correlations with human judgements in comparison with BLEU. ChrF++ relies on character-level matches and computes the harmonic mean of character-level n-grams precision and recall between reference and predictions. To compute ChrF++, we set word order as 1 and character n-gram as 4. Following BLI literature, we also report P@1 scores for all our experiments.

**Human Evaluation**: We extend the IDDT experiment (refer Section 4.2) and additionally perform human evaluations on two language pairs,

| Language | Transliteration (in %) | Word Length | | | |
|---|---|---|---|---|---|
| | | 1 | 2 | 3 | 4 |
| Hindi | 1.4 | 52.5 | 39.4 | 6.5 | 0.9 |
| Tamil | 0.01 | 48.1 | 42.7 | 7.8 | 1 |
| Gujarati | 0.75 | 53.9 | 37.9 | 6.5 | 1.2 |
| Odia | 0.3 | 51 | 41.4 | 6.2 | 1 |
| Kannada | 0.05 | 56.8 | 35.5 | 0.9 | 6.6 |
| Marathi | 0.59 | 53.2 | 39.6 | 6.1 | 0.9 |

Table 1: Frequency distribution of word length and percentage of transliterated dictionary pairs in the dataset for different languages.

*viz.*, Punjabi and Malayalam, a north Indian and a south Indian language, respectively. We provide the top three predictions generated by our model to the expert annotators and ask them to select valid translation(s). If there are multiple translations for a phrase, we ask annotators to provide a preference order for the translations. The test set comprises 200 instances from each of the administrative and biotechnology domains across all two language pairs. We use recall@1 and recall@3 metrics to evaluate our predictions.

### 4.4 Datasets

Our dictionary translation dataset comprises of expert-curated dictionaries provided by a government agency, Commission for Scientific and Technical Terminologies (CSTT)[3]. This is an apex Indian agency to prepare multi-lingual domain glossaries in 22 official Indian languages including English. We created the dataset by extracting the text from the dictionaries available as a digitized PDF. The dictionary pairs consists of 6583, 3678, and 1809 parallel translation pairs across the administrative, chemistry, and biotechnology domains respectively, replicated across each of the six language pairs. These dictionaries include terms ranging from single-word to 4-word phrases, while majorly being single-word.

### 4.5 Dataset Distribution and Transliteration

We present distribution of word length for the corresponding language in Table 1. Note that majority of the target language dictionary phrases are single-word, however, significant two length words are present in the target dictionary. The complete dataset is present at github link

[3]https://www.csttpublication.mhrd.gov.in

In Table 1, we also present the percentage of dictionary pairs which are transliterations of source term. We use Indic-Xlit (Madhani et al., 2022) library to transliterate the source language word into target language. We observe that Tamil has the least amount of transliteration from English amounting to merely 0.01%. Only one word 'visa' is transliterated into Tamil as விசா. After Tamil, Kannada has the least amount of transliterations followed by Odia, Marathi and Gujarati. Hindi lexicon contains most transliterations of English among all 6 languages.

### 4.6 Baselines

We mention implementation details for LexGen and other baselines in Appendix A. In LexGen, due to introduction of DR layer in each decoder block, we have an additional 12 million parameters. Thus, overall count of parameters for our LexGen is $\approx 492M$. All results reported in the paper are for a single run.

1. **Base Transformer (Base)**: We use a pre-trained Transformer model trained on the Samanantar dataset (Ramesh et al., 2022) consisting of 49 million parallel sentences from English to 11 Indic languages. The pre-trained model is further fine-tuned on the training set which updates all model parameters for the dictionary generation task. The transformer model consists of 16 attention heads for both encoder and decoder, 1536 embedding dimensions, and 4096 feed-forward embedding dimensions. Training takes approximately 20 minutes for 1 epoch on a single Nvidia A6000.

2. **NLLB (Team, 2024)**: We use a pre-trained NLLB-200 NMT model which is trained using sparsely gated mixture of experts architecture. It provides separate gates for each language thus capturing language-specific nuances and helps decrease interference (in the learned weights) between unrelated languages. We perform full parameter fine-tuning using our dictionary translation dataset. The experts are then combined using a gating mechanism to produce the final translation. We use the NLLB model distilled 600M variant from HuggingFace[4] to fine-tune with our training dataset.

3. **BLICEr** (Li et al., 2022) is a lexicon induction approach that initially creates a seed set of contrastive word similarity dataset across language pairs. Thereafter, it finetunes a multi-lingual pre-trained language model (mPLM) in a cross encoder manner to predict the similarity scores. During inference, it combines the similarity scores from the original cross-lingual space and mPLM and selects the target phrase having the highest similarity score with the source phrase.

[4]https://huggingface.co/facebook/nllb-200-distilled-600M

| Test Dataset | Model | Gujarati | Hindi | Kannada | Marathi | Odia | Tamil | Avg ChrF | Avg P@1 |
|---|---|---|---|---|---|---|---|---|---|
| Administrative | ICL | 4.59 | 54.13 | 4.59 | 40.49 | 3.6 | 7.93 | 19.22 | 0.10 |
| | FT LLM | 5.37 | 58.37 | 5.75 | 44.56 | 4.11 | 8.12 | 21.04 | 0.10 |
| | BLICEr | 41.23 | 54.21 | 45.86 | 40.72 | 34.70 | 40.49 | 42.98 | 0.17 |
| | NLLB | 47.72 | 57.44 | 50.97 | 50.12 | 47.09 | 48.47 | 50.30 | 0.25 |
| | Base | 54.74 | 68.11 | 52.64 | **56.20** | 52.70 | 48.97 | 55.56 | 0.33 |
| | LexGen | **55.39** | **68.19** | **54.25** | 55.04 | **54.23** | **53.97** | **56.84** | **0.34** |
| Biotechnology | ICL | 3.61 | 41.85 | 4.31 | 41.21 | 2.81 | 9.12 | 17.15 | 0.04 |
| | FT LLM | 4.27 | 43.71 | 5.46 | 43.99 | 3.42 | 9.33 | 18.36 | 0.05 |
| | BLICEr | 49.95 | 54.08 | 48.30 | 49.81 | 40.38 | 39.83 | 47.06 | 0.08 |
| | NLLB | 42.09 | 52.06 | 36.77 | 50.98 | 42.64 | 31.38 | 42.65 | 0.11 |
| | Base | 64.19 | 63.91 | 61.69 | 63.70 | 59.07 | 53.75 | 61.05 | 0.25 |
| | LexGen | **68.88** | **69.50** | **64.78** | **69.48** | **63.04** | **53.97** | **64.94** | **0.29** |
| Chemistry | ICL | 3.3 | 43.36 | 4.23 | 43.77 | 2.72 | 8.31 | 17.61 | 0.07 |
| | FT LLM | 4.16 | 44.6 | 4.99 | 46.18 | 3.17 | 8.45 | 18.59 | 0.07 |
| | BLICEr | 44.98 | 54.73 | 41.16 | 45.05 | 35.18 | 31.35 | 42.07 | 0.09 |
| | NLLB | 39.56 | 47.50 | 37.70 | 39.11 | 37.33 | 27.82 | 38.17 | 0.11 |
| | Base | 55.98 | 57.19 | 55.37 | 55.87 | 52.62 | **44.55** | 53.60 | 0.24 |
| | LexGen | **58.66** | **59.12** | **60.07** | **60.02** | **54.71** | 44.06 | **56.11** | **0.29** |

Table 2: ChrF scores on administrative, biotechnology, and chemistry for in-domain in-language (IDST). ICL and FT refers to in-context learning and fine-tuning over Openhathi (Hindi fine-tuned version over Llama2). We also report average precision@1 scores for each model across all languages in Table 8.

4. **In-context learning (ICL)**: Li et al. (2023) demonstrated the efficacy of utilizing in-context examples generated through nearest neighbor search using text-to-text LLMs for the task of dictionary generation. For a source phrase, the in-context examples are chosen by picking the top-5 nearest source phrases from the train set based on cosine similarity. We use OpenHathi[5] which is a Llama-based model (Touvron et al., 2023) with vocabulary extension and continual pre-training over Hindi corpora. The chosen in-context examples from our dataset are used to generate dictionary translation for the test instances.

5. **Fine-tuning LLM (FT LLM)**: We performed QLoRA (Dettmers et al., 2023) fine-tuning with our training dataset over the OpenHathi model. Prompt and fine-tuning details are present in Section A.

[5] https://www.sarvam.ai/blog/announcing-openhathi-series

## 5 Results

### 5.1 In-domain Evaluation (IDST)

Table 2 shows the performance of LexGen and baselines on in-domain same-language test set. The base Transformer model, pre-trained on a generic Indic language corpus, exhibits notable proficiency in the administrative domain but falls short on the biotechnology and chemistry domains. The underlying reason could be that the base Transformer model is pre-trained on a general news and administrative corpus enabling the model to learn the commonness in vocabulary and inter-relatedness of languages for administrative phrases. Due to the presence of the DR layer, LexGen can effectively learn the differential representations varying across the training domains. The average test-set phrase overlap across these three domains is approximately 7%, leading to weaker performance by the Transformer baseline without a DR layer when compared to LexGen. Furthermore, roughly 10% overlap exists among the training instances between the biotechnology and chemistry domains. This overlap results in a wider performance gap between LexGen and the base Transformer baseline, especially compared to the administrative domain.

In-context learning and QLoRA fine-tuning over Hindi fine-tuned Llama2 model yield sub-optimal results on all languages except Hindi and Marathi. Continual pre-training on a substantial Hindi corpus enhances performance on Marathi, likely due to the significant linguistic overlap between the two languages. Conversely, other languages demonstrate poorer results as the model has not been exposed to sufficient data from these languages during the pre-training phase. Notably, fine-tuning yields better ChrF++ scores on all languages including Hindi and Marathi. Therefore, we omit reporting ICL, FT LLM for the remaining experiments. BLI baseline, BLICEr, reports poorer ChrF++ and P@1 scores primarily due to its inability to handle multi-word phrases. BLI approaches align word and phrasal embeddings while training, however unseen words and phrases during testing cannot be mapped with the corresponding target phrases resulting in low scores on IDST setting. Hence, for further zero-shot IDDT experiments, we omit reporting BLICEr scores. Additionally, we present language-wise precision results in Table 8.

| Test Dataset | Model | Kannada | Marathi | Odia | Average |
|---|---|---|---|---|---|
| Administrative | NLLB | 48.36 | 45.22 | 43.88 | 45.82 |
| | Base | 45.99 | 51.05 | 45.02 | 47.35 |
| | LexGen | **52.33** | **52.37** | **48.66** | **51.12** |
| Biotechnology | NLLB | 24.76 | 39.95 | 24.50 | 29.74 |
| | Base | 52.10 | 59.80 | 55.90 | 55.93 |
| | LexGen | **58.30** | **63.10** | **58.20** | **59.87** |
| Chemistry | NLLB | 24.03 | 36.40 | 24.41 | 28.28 |
| | Base | 36.90 | 48.50 | 44.30 | 43.23 |
| | LexGen | **41.20** | **51.80** | **46.90** | **46.63** |

Table 3: ChrF++ scores on administrative, biotechnology, and chemistry for unseen languages, namely, Kannada, Marathi, and Odia (in-domain different-target - IDDT experiment).

### 5.2 Zero-shot Cross-lingual Evaluation (IDDT)

In Table 3, we present results in a zero-shot setting, testing on the same domain but a different target language. Our goal is to analyze the language transfer capabilities of our approach without provisioning for any additional language routing mechanism. The aim is to leverage inter-relatedness among Indic languages and enable information sharing across languages. LexGen achieves superior and statistically significant results compared to the base Transformer baseline on all three domains on unseen languages. Due to its training corpus being closely aligned with the administrative domain, NLLB ChrF++ scores have a smaller gap with respect to LexGen in the administrative domain in comparison to the biotechnology and chemistry domains.

| Language | Intersection | Non-intersection | % of non-transliterations |
|---|---|---|---|
| Hindi | 0.51 | 0.27 | 0.59 |
| Kannada | 0.15 | 0.13 | 0.96 |
| Gujarati | 0.25 | 0.18 | 0.31 |
| Marathi | 0.28 | 0.24 | 0.62 |

Table 4: P@1 results for out-domain-same-target experiments averaged across all domains for LexGen. Intersection column refers to scores for phrases that has source-side overlap between train and test sets. Last column refers to the percentage of non-transliterations for non-intersecting predictions.

### 5.3 Domain Generalization (DDST)

In this experiment, we evaluated the out-of-domain prediction ability of LexGen by leveraging $W_{\text{shared}}$ and multi-lingual pre-trained base model. We intend to assess the out-of-domain generalization ability of our framework in predicting unseen out-of-domain instances. Our test set consists of 500 instances from unseen domains namely broadcasting, civil engineering, computer engineering, geography, and psychology in Hindi, Kannada, Gujarati and Marathi language.

To ensure that test set does not have a source-side overlap with the train set, we evaluated predictions on overlapping and non-overlapping words between train and test set. To perform this experiment, we find the source side (English) overlap between train and test sets. First, we created a unique set of words from both sets by splitting each phrase by space. Second, the overlap and non-overlap of these sets form our intersection and non-intersection sets respectively. Third, we perform alignment of source and target phrases using Awesome Align (Dou and Neubig, 2021) to align source and target words. Finally, we compare the target-side prediction with the ground truth for each word for the intersecting and non-intersecting sets.

Table 4 presents averaged P@1 results for the overlap and non-overlapping words between train and test sets across all domains. To confirm whether predictions on non-intersecting words are not mere transliterations, we also check the percentage of non-transliterated words among the non-intersected predictions. The results suggests that those words which are present both in train and test sets have higher chances of correct predictions for unseen domains. However, unique words occurring only in the test set have non-zero scores. This implies that introduction of $W_{\text{shared}}$ and a pre-trained multi-lingual model is beneficial for predicting out of domain words. It confirms that predicted translations are not merely transliterations of English words. For Gujarati, nearly 2/3rd of predicted translations are transliterations. Conversely for Hindi, Kannada and Gujarati, merely 1/3rd of predicted translations are transliterations.

| Test Dataset | Model | Average |
|---|---|---|
| Administrative | Shared gate layer only | 56.65 |
| | DR layer after CAN | 49.07 |
| | DR layer after SAN | **56.84** |
| Biotechnology | Shared gate layer only | 61.72 |
| | DR layer after CAN | 61.97 |
| | DR layer after SAN | **64.94** |
| Chemistry | Shared gate layer only | 54.97 |
| | DR layer after CAN | 54.87 |
| | DR layer after SAN | **56.11** |

Table 5: Average ChrF++ scores for administrative, biotechnology, and chemistry domains for IDST with different DR scenarios across 6 languages. CAN and SAN refers to cross attention and self attention layers in the decoder blocks respectively.

### 5.4 Impact of DR layer

The position where the DR layer is inserted impacts model performance. In Table 5, we present results using different positions of the DR layer and without $W_{\text{dom}}$. The results are reported for the IDST setting, with each row representing the average numbers across six language pairs. When $g(z_l)$ is set to 0 and only $W_{\text{shared}}$ is activated (refer to Eq 1), the results are inferior compared to when the full DR layer is activated following the self-attention network layer. The introduction of the DR layer after CAN results in slightly lower scores than with shared gate layer scenario for the administrative and chemistry domains. Placing the DR layer after CAN results in degradation of the results. We hypothesize this could be due to DR's role being most effective in identifying domain knowledge right before cross-attention with the encoder; using it after CAN could erase useful encoder-specific information.

| | Punjabi | | Malayalam | |
|---|---|---|---|---|
| | LexGen | Base | LexGen | Base |
| R@1 | **0.52** | 0.51 | 0.38 | 0.38 |
| R@3 | **0.97** | 0.95 | **0.71** | 0.68 |

Table 6: Human post-hoc evaluation on same-domain but different target language in a zero-shot setting. Recall at 1 and 3 for two languages.

### 5.5 Human post-hoc evaluation

We augment the IDDT experiment (refer Section 4.2) to include two additional languages, namely Punjabi and Malayalam, and conduct a post-hoc manual inspection of the generated predictions shown in Table 6. Similar to the IDDT experiment, our training dataset comprises dictionary translations in English-Hindi, English-Gujarati, and English-Tamil.

For the assessment, we follow the human evaluation metric as explained in Section 4.3. LexGen reports better R@3 for Punjabi which can be ascribed to the language's shared features with other languages. Punjabi has a considerably higher linguistic overlap with Hindi (which is a part of the training set), resulting in significantly better R@1 and R@3 when compared to Malayalam. It is noteworthy that Malayalam demonstrates good scores on R@1 and R@3 despite having a different script and different language. This can be primarily attributed to the ability of leveraging shared language features in the pre-trained model.

Table 7 in Appendix lists few example prompts used for the ICL experiment. Through comprehensive analysis, we found the first prompt most effective. Note that we used 5-shot examples in the actual prompt during inference.

## 6 Conclusion

We present LexGen, an approach to lexicon generation for six Indian languages across diverse domains using pre-trained MNMT models. LexGen effectively combines both domain-specific and domain-generic layers through learnable gates to generate accurate and domain-relevant dictionary words. Our experimental results, conducted in both zero-shot and few-shot scenarios, demonstrate the ability of our model to generalize to unseen domains and languages. Further, we also release a benchmark dataset spanning eight domains and six languages consisting of more than 75K translation pairs for further research in domain-specific lexicon generation, particularly for Indic languages.

## 7 Limitations

Firstly, we have limited our experiments to only a single MNMT model. We did not explore the impact of different NMT models for each language and adding DR layers within each model. As part of future work, multiple NMT models for each language can be explored. Secondly, our experimental scope is confined to Indic languages, and we did not include a larger set of (non-Indic) languages.

## Acknowledgements

We acknowledge BharatGen and IIT Bombay for providing resources and support for the project. Karthika is supported by the PhD fellowship from the TCS Research Foundation. We also extend our appreciation to the reviewers for their valuable feedback.

## A Implementation Details

**Base Transformer and LexGen.** We implement Base Transformer and LexGen using fairseq toolkit (Ott et al., 2019) v0.12 over Transformer model for all our experiments. We utilize a multilingual pre-trained checkpoint from (Ramesh et al., 2022) consisting of $\approx 480M$ parameters to initialize the model. The reported results are for a single run. The transformer model consists of 16 attention heads for both encoder and decoder, 1536 embedding dimensions, and 4096 feed-forward embedding dimensions. The optimizer employed is Adam (Kingma and Ba, 2014) with label smoothing of $0.1$, the learning rate is set to $0.0001$ with $4000$ warm-up steps, the probability dropout is set to $0.2$, maximum token length to $1024$ and the maximum number of updates to $200,000$ with an early stopping patience of 5 epochs. The beam size for all experiments and baselines is set to 5. Training takes approximately 20 minutes for 1 epoch on a single Nvidia A6000.

**LexGen** Due to introduction of DR layer in each decoder block, we have an additional 12 million parameters. Thus, overall count of parameters for our model is $\approx 492M$.

**NLLB.** We use the NLLB model distilled 600M variant[6] from HuggingFace to fine-tune with our training dataset.

**BLICEr.** Following the work from (Li et al., 2022), BLICEr utilizes FastText monolingual word embeddings to generate initial cross-lingual word embeddings. This process follows the Contrastive Bilingual Language Identification (ContrastiveBLI) framework, specifically employing either the C1 or C2 stage and adopting hyperparameter settings from the original study. For similarity score prediction, BLICEr finetunes the XLM-R(large) model from the sentence-transformers module in a cross-encoder manner. BLICEr does not involve text generation. Instead, it calculates similarity scores between the input source word/phrase and all ground-truth words/phrases on the target side. The target word/phrase with the highest similarity score is then selected as the predicted translation. In BLICEr, we replaced train-test set used in BLICEr with our train-test set, keeping the experimental setup mostly unchanged.

**LLaMA Based In-Context Learning and Fine-Tuning.** We used the pre-trained checkpoint of LLaMA2-7B available under the HuggingFace model hub[7]. The experiments were performed in two stages. The first stage was in-context learning using examples derived from nearest-neighbour search. In the second stage, we fine-tuned LLMs. We used QLoRA (Dettmers et al., 2023) for efficient finetuning of LLMs and the implementation of QLoRA is done using the peft module of Hugging Face. In LoraConfig we set the lora_alpha to 16, lora_dropout to 0.1, and r to 64. In BitsAndBytesConfig we set bnb_4bit_compute_dtype to bfloat16. The optimizer used is AdamW (Loshchilov and Hutter, 2017), the learning rate is set to 0.00001 and the weight decay is set to 0.1. Training takes approximately 1 hour for 1 epoch on a single Nvidia A100.

## B Human Annotation Details

For the post-hoc human evaluation, we used beam decoding method to get the top-3 predictions and provided them to experts to rank the translations. We recruited expert linguistic translators who are domain experts and worked as part of at least 3 book translation project. The reviewers were asked to check if the predictions are correct, and if multiple predictions are correct, they were asked to rate them in the order of the best predictions (for reporting R@1 and R@3). We would also like to emphasize that the ground truth for the terms for which the predictions were made was not available. Hence, the annotator review is purely based on their knowledge of the language and not based on any comparison with ground truth.

We provided source phrases in Microsoft Excel and asked translators to fill in the corresponding translations. All the translators were duly compensated as per industry standards. The following instructions were given to the translators: "For a given source phrase, please translate in the corresponding target language and domain using your domain and linguistic expertise. Please prioritize quality over speed of translation to ensure accuracy. We do not collect personal information and your translations will remain anonymous. By participating, you agree to grant us a non-exclusive, royalty-free license to use the translations for research purposes."

[6] https://huggingface.co/facebook/nllb-200-distilled-600M

[7] https://huggingface.co/meta-llama/Llama-2-7b-hf

1. The $L_a$ word $w_a^1$ in $L_b$ is $w_b^1$. The $L_a$ word $w_a^2$ in $L_b$ is $w_b^2$. The $L_a$ word $w_a^3$ in $L_b$ is

2. The translation of the word $w_a^1$ from $L_a$ to $L_b$ under physics domain is $w_b^1$.
The translation of the word $w_a^2$ from $L_a$ to $L_b$ under physics domain is $w_b^2$.
The translation of the word $w_a^3$ from $L_a$ to $L_b$ under physics domain is

Table 7: Example of prompts for the In-Context-Learning (ICL) experiments. Here, we assume the number of few-shot examples to be two. $L_a$ and $L_b$ correspond to the source and the target language respectively. ($w_a^1$, $w_b^1$) and ($w_a^2$, $w_b^2$) are the few-shot examples. Note that we used 5-shot examples in the actual prompt during inference.

| Test Dataset | Model | Gujarati | Hindi | Kannada | Marathi | Odia | Tamil | Average |
|---|---|---|---|---|---|---|---|---|
| | ICL | 0.0 | 0.38 | 0.0 | 0.21 | 0.0 | 0.0 | 0.10 |
| | FT LLM | 0.0 | 0.39 | 0.0 | 0.23 | 0.0 | 0.0 | 0.10 |
| Administrative | BLICEr | 0.18 | 0.26 | 0.19 | 0.16 | 0.11 | 0.13 | 0.17 |
| | NLLB | 0.28 | 0.37 | 0.15 | 0.29 | 0.22 | 0.23 | 0.25 |
| | Base | 0.34 | **0.48** | 0.31 | **0.34** | 0.27 | 0.25 | 0.33 |
| | LexGen | **0.34** | 0.46 | **0.32** | 0.33 | **0.28** | **0.29** | **0.34** |
| | ICL | 0.0 | 0.11 | 0.0 | 0.14 | 0.0 | 0.01 | 0.04 |
| | FT LLM | 0.0 | 0.12 | 0.0 | 0.14 | 0.0 | 0.01 | 0.05 |
| Biotechnology | BLICEr | 0.06 | 0.15 | 0.09 | 0.10 | 0.07 | 0.04 | 0.08 |
| | NLLB | 0.10 | 0.18 | 0.08 | 0.18 | 0.08 | 0.03 | 0.11 |
| | Base | 0.26 | 0.33 | 0.25 | 0.30 | 0.23 | 0.14 | 0.25 |
| | LexGen | **0.31** | **0.34** | **0.30** | **0.33** | **0.26** | **0.22** | **0.29** |
| | ICL | 0.0 | 0.19 | 0.0 | 0.21 | 0.0 | 0.0 | 0.07 |
| | FT LLM | 0.0 | 0.19 | 0.0 | 0.23 | 0.0 | 0.0 | 0.07 |
| Chemistry | BLICEr | 0.11 | 0.15 | 0.09 | 0.09 | 0.07 | 0.05 | 0.09 |
| | NLLB | 0.11 | 0.19 | 0.11 | 0.12 | 0.09 | 0.04 | 0.11 |
| | Base | 0.28 | 0.29 | 0.26 | 0.30 | 0.21 | 0.12 | 0.24 |
| | LexGen | **0.32** | **0.30** | **0.32** | **0.33** | **0.26** | **0.19** | **0.29** |

Table 8: P@1 scores on administrative, biotechnology, and chemistry for in-domain in-language (IDST) setting.

| Target Lang | English | Ground Truth | Base | LexGen |
|---|---|---|---|---|
| | toxic genomics | आविष–संजीनिकी | विषाक्त जीनोमिक्स | आविषी संजीनिकी |
| | phagocytosis | भक्षकोशिकता, फैगोसाइटोसिस | फैगोसाइटोसिस | भक्षकाणु |
| Hindi | streamline | सुप्रवाही बनाना | स्ट्रीमलाइन | सुव्यवस्थित करना |
| | stereospecific catalyst | त्रिविमविशिष्ट उत्प्रेरक | स्टीरियोस्पेसिफिक उत्प्रेरक | त्रिविमी विशिष्ट उत्प्रेरक |
| | Isomorphism | समाकृतिकता | समरूपता | समाकृतिकता |
| | non ideal gas | अनादर्श गैस | अनादर्श गैस | अनादर्श गैस |
| | Precooling | ಪೂರ್ವಶೀತಲನ | ಪೂರ್ವಊತನ | ಪೂರ್ವಊಷ್ಮಣ |
| Kannada | essential | ಸಾರಭೂತ,ಅನಿವಾರ್ಯ , ಸಗಂಧ | ಅತ್ಯವಶ್ಯಕ | ಅನಿವಾರ್ಯ |
| | Output | ನಿರ್ಗಮ | ಔಟ್ಪುಟ್ | ನಿರ್ಗಮ |
| | revolving axis | ಪರಿಕ್ರಾಮೀ ಅಕ್ಷ | ಘೂರ್ಣನ ಅಕ್ಷ | ಪರಿಕ್ರಾಮೀ ಅಕ್ಷ |
| | gene probe | जीन संपरीक्षक | जीन जांच | जीन संपरीक्षक |
| | cytotoxic | कोशिकाआविषी | साइटोटॉक्सिक | कोशिकाआविषण |
| Hindi | metallo enzyme | धातु एन्जाइम | धात्विक एंजाइम | धातु एन्जाइम |
| | acquired immunity | उपार्जित प्रतिरक्षा | अर्जित प्रतिरक्षा | उपार्जित प्रतिरक्षा |
| | genomics | संजीनिकी | जीनोमिक्स | संजीनिकी |

Table 9: Predicted samples for baseline (Base Transformer) and LexGen for the biotechnology and chemistry domain. The base transformer without DR layer transliterate in most of the cases while LexGen aligns more closer to the ground truth.